\documentclass{article}

\usepackage{arxiv}

\usepackage[utf8]{inputenc} 
\usepackage[T1]{fontenc}    
\usepackage{hyperref}       
\usepackage{url}            
\usepackage{booktabs}       
\usepackage{amsfonts}       
\usepackage{nicefrac}       
\usepackage{microtype}      
\usepackage{lipsum}		
\usepackage{graphicx}
\usepackage{natbib}
\usepackage{doi}
\usepackage{amsmath,amssymb}
\usepackage{multirow}
\usepackage{subcaption}
\usepackage{tcolorbox}

\title{Explicit Context Integrated Recurrent Neural Network for Sensor Data Applications}

\author{ \href{https://orcid.org/0000-0002-6756-8157}{\includegraphics[scale=0.06]{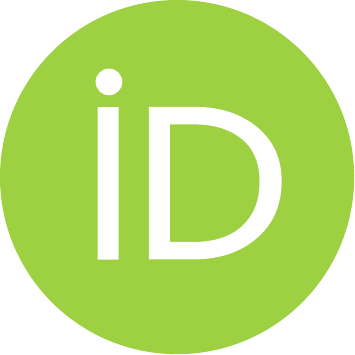}\hspace{1mm}Rashmi Dutta Baruah}\thanks{Department of Computer Science and Engineering, IIT Guwahati, Guwahati 781039, Assam, India} \\
	Department of Telematic Engineering\\
	Universidad Carloss III de Madrid\\
        Avda. de la Universidad, 30\\
	Lagan\`es, Madrid, 28911, Spain\\
	\texttt{rdutta@it.uc3m.es} \\
	\And
	\href{https://orcid.org/0000-0000-0000-0000}{\includegraphics[scale=0.06]{orcid.pdf}\hspace{1mm}Mario Mu\~noz Organero} \\
	Department of Telematic Engineering\\
	Universidad Carloss III de Madrid\\
        Avda. de la Universidad, 30\\
	Lagan\`es, Madrid, 28911, Spain\\
	\texttt{mario.munoz@it.uc3.es} \\
}

\renewcommand{\shorttitle}{Context Integrated Recurrent Neural Network}

\begin{document}
\maketitle
\begin{abstract}
 The development and progress in sensor, communication and computing technologies have led to data rich environments. In such environments, data can easily be acquired not only from the monitored entities but also from the surroundings where the entity is operating. The additional data that are available from the problem domain, which cannot be used independently for learning models, constitute \textit{explicit} context. Such context, if taken into account while learning, can potentially improve the performance of predictive models. Typically, the data from various sensors are present in the form of time series. Recurrent Neural Networks (RNNs) are preferred for such data as it can inherently handle temporal context. However, the conventional RNN models such as Elman RNN, Long Short-Term Memory (LSTM) and Gated Recurrent Unit (GRU) in their present form do not provide any mechanism to  integrate explicit contexts. In this paper, we propose a Context Integrated RNN (CiRNN) that enables integrating explicit contexts represented in the form of contextual features. In CiRNN, the network weights are influenced by contextual features in such a way that the primary input features which are more relevant to a given context are given more importance. To show the efficacy of CiRNN, we selected an application domain, engine health prognostics, which captures data from various sensors and where contextual information is available. We used the NASA Turbofan Engine Degradation Simulation dataset for estimating Remaining Useful Life (RUL) as it provides contextual information. We compared CiRNN with baseline models as well as the state-of-the-art methods. The experimental results show an improvement of 39\%  and 87\% respectively, over state-of-the art models, when performance is measured with RMSE and score from an asymmetric scoring function. The latter measure is specific to the task of RUL estimation. 
\end{abstract}


\section{Introduction}
\label{sec:introduction}
The Internet of Things (IoT) has led to many 'smart' innovations such as smart homes, smart cities, smart factories, smart agriculture, to name a few. The primary contributors to IoT enabled smart creations are advanced sensor, communication, and computing technologies. The sensors play a crucial role as IoT devices depend on them to get the data from the monitored entities. The monitored entity can range from a person (for example, to get health parameters), environment (to get air, water, or soil parameters), to a large industrial setup (to get parameters associated to various processes or equipment). However, acquiring and storing data is not sufficient, the data needs to be analysed and interpreted to extract knowledge, which can further be used for decision making and automation at various levels. Machine Learning (ML) offers a way to achieve this through predictive analytics. Particularly, Deep Learning (DL) has gained a lot of attention due to huge amount of data generated by smart environments (homes, cities, industries etc.) \citep{Chen2014}, \citep{Naj2014}. Typically, the data received through sensors from such domains are time series data. Recurrent Neural Networks (RNNs) can be a preferable approach as they inherently capture temporal context available in the time series or sequential data \cite{KARIM2019}, \cite{Gers2001}, \cite{Malhotra2015}.

One of the early and widely known RNNs, Elman network \citep{ELMAN1990}, introduced the notion of context neurons that function as network’s memory and allow implicitly representing the temporal context of the input data. The input layer is augmented with a context layer. The context units are initialized to a specific value. At a particular time step $t$, the hidden units are activated by both input layer units and context layer units. The hidden units with feed forward connections, in turn, activates the output layer units. The hidden units also activate the context unit through feed-back connections which have unit weight. Thus, in the next time step $t+1$, the context units have the hidden unit values of previous time step $t$. The hidden unit patterns is the context that is saved in the context units. The hidden units map both the input and the previous state to desired output. Thus, the internal representations developed in the process are sensitive to the temporal context \citep{ELMAN1990}. In this paper, we treat the notion of context in a different way, and the focus is not on the temporal context or the contexts that are generated within the network from input and/or output signals as we already have various forms of RNNs that elegantly capture them. 

The notions of context, context-aware, context-dependent, and context-sensitive are widely used across various domains. Several definitions of these terms, particularly for the term context, have appeared in the literature. During the early 90’s these concepts gained huge popularity within the research community of pervasive and ubiquitous computing, which evolved from mobile and distributed computing. As defined by Abowd et al.\citep{Abowd1999}, context is any information that can be used to characterise the situation of an entity. An entity is a person, place, or object that is considered relevant to the interaction between a user and an application, including the user and applications themselves. From machine learning perspective, we are concerned with context that is available as data from the problem domain but not necessarily independently used for learning. The context data may influence the decision by improving the model but may not be involved directly in learning. For example, in medical diagnosis, the patient’s gender, age, and weight are often available, which can be considered as contextual data. For clarity, we adopt the distinction between \textit{primary} features and \textit{contextual} features as described in \citep{Turney1996}. Primary features are useful for machine learning task (classification/regression) when considered in isolation, irrespective of other features. Contextual features are not useful in isolation, but can be useful when combined with primary features. For example, for a gas turbine engine diagnosis problem, thrust, temperature, and pressure are primary features, whereas the weather conditions such as ambient temperature and humidity can be considered as contextual features. We also distinguish between \textit{explicit} and \textit{implicit} context. The context that can be captured from the environment where primary data is acquired is considered as explicit. The above-mentioned examples of medical diagnosis and gas turbine engine diagnosis have explicit contexts represented by contextual features. On the other hand, implicit context is present in the primary data itself. For example, for image classification, many approaches consider the neighbouring pixels as context while processing a particular pixel. Similarly, in Natural Language Processing (NLP) task, the neighbouring words can be considered as context. Finally, the temporal sequence of the instances in a data set is another example of implicit context. For simplicity, in rest of the paper, we will be referring 'explicit context' as 'context'. 

In several problem domains, data from operating environment is often captured along with the data from monitored system (or entity), in other words, the contextual information is available. Yet, during learning, the focus is on the internal state data of the monitored entity and the external data from the operating environment is not fully considered or even ignored. As mentioned earlier RNNs (including Long Short Term Memory-LSTM and Gated Recurrent Unit-GRU) can handle the temporal context very well. However, there is no explicit way of leveraging the contextual information. Usually, the feature space comprising of primary features is expanded with contextual features, and the models treat them in the same manner as primary features. 

In this paper, we propose an approach that integrates explicit contexts to RNN and enhances its potentials. We refer to the the resulting RNN as Context Integrated RNN (CiRNN) which takes into account implicit temporal context as well as explicit context \citep{Baruah2023}. In the proposed approach, the contextual features are used to weight the primary features such that features that are more relevant in a given context are assigned more importance. This is also termed as \textit{contextual weighting} \citep{Turney1996}. This work draws inspiration from the work of Ciskowski and Rafajlowicz \citep{Ciskowski2004} where they introduced Context-Dependent Feed-Forward Neural Networks. We demonstrate the efficacy of CiRNN by applying it to the domain of predictive maintenance where contextual information is available through various sensors.  

The rest of the paper is organised as follows. In the next section we briefly discuss methods where context is added to RNN in various ways. In section 3, we discuss the architecture and learning of the proposed context integrated
RNN with GRU. Section 4, discusses the experiments and
results achieved. Finally, section 5 concludes the paper.

\section{Related Work}

In the past decade, many researchers have considered integrating context to RNNs. These methods largely appear in the literature of NLP and recommender systems.  In this section, we briefly discuss various RNN architectures that leverage context information for better performance in a given task. 

Mikolov and Zweig \citep{Tomas2012} proposed a context dependent recurrent neural network language model (CDRNNLM) that extends the basic recurrent neural network language model (RNNLM) by introducing context. Context is a real valued vector that is computed based on the sentence history using Latent Dirichlet Allocation (LDA). The architecture introduces a feature layer in addition to the input, output, and hidden layer. The feature layer with associated weights is connected to both hidden and output layers. It provides the context i.e. complementary information to the input word vector, for example, features representing topic information. The conventional approach is to partition the data and to develop multiple topic specific models. However, through CDRNNLM a topic-conditioned RNNLM is attained that avoids the data partitioning. They demonstrated that CDRNNLM improves the performance over state-of-the art methods using benchmark dataset.

In \citep{wang-cho-2016}, a method is proposed that incorporates corpus-level discourse information into language modelling and the model is referred to as larger-context language model. The effect of context is modeled by a conditional probability where the probability of the next words in a given sentence is determined using previous words in the same sentence, and a context sentence that consists of one or more sentences of arbitrary length. To realize this, a conditional LSTM is proposed which is implemented in two ways, early fusion and late fusion. In early fusion, the context vector representation of preceding sentences is simply considered as an input at every time step. Late fusion tries to maintain separately the dependencies within the sentence being modelled (intra-sentence dependencies) and those between the preceding sentences and the current sentence (inter-sentence dependencies). The existing LSTM memory cell $c_t$ models the intra-sentence dependency. The inter-sentence dependency is modeled through the interaction between $c_t$ and the context vector. First, a controlled context vector $r_t$ is determined using $c_t$ and the context vector. The vector $r_t$ represents the amount of influence of context or preceding sentences. Finally, the controlled context vector is used to compute the output of the LSTM layer. The experimental results showed that later fusion outperformed early fusion. 

A Language Model that uses a modified version of the standard bidirectional LSTM called contextual bidirectional LSTM (cBLSTM) is proposed in \citep{mousa-schuller-2017} for sentiment analysis. cBLSTM predicts a word given the full left and right context while excluding the predicted word itself from the conditional dependence. This is achieved by introducing a forward and a backward sub-layer. The encoded input consists of sentence start symbol and all the words before the sentence end symbol. It is given to the forward sub-layer and the forward hidden states predict the words between start and the end symbol. The backward sub-layer does the reverse operation. When the two sub-layers are interleaved it results in prediction of any target word using the full left and right contexts. The obtained results on benchmark dataset show that cBLSTM language model outperforms both LSTM and BLSTM based models. In \citep{Smirnova2017}, a family of Contextual RNNs (CRNNs) is presented that leverages contextual information for item sequence modeling for next item prediction in a recommender system. Apart from order of items, the model considers other available information about user item interaction such as type of the interaction, the time gaps between events and time of interaction as context. The two proposed CRNN architectures are compared with non-contextual models and the results show that CRNNs significantly outperforms all of the other methods on two e-commerce datasets. 

Tran et al. \citep{tran-etal-2018-context} proposed
a Context-dependent Additive Recurrent Neural Network
(CARNN) for the problem of sequence mapping in NLP. The learning process considers the primary source of information as the given text, for example,
the history of utterance in a dialog system or the sequence of words in language modeling. The previous utterance in dialog system or the discourse information in language modeling is taken as context. The CARNN units consist of two gates (update gate and reset gate) that regulate the information from the input. The gates are functions that depend on global context, word embedding (input), and previous hidden vector. In \citep{Wenke2019}, a RNN architecture named as Contextual RNN is discussed where contextual information is used to parameterize the initial hidden state. Typically, the hidden state of RNN is initialised to a default constant, most commonly to zero. In this paper, the initial hidden state is conditioned on contextual information. The contextual information can be acquired from the input sequence or from a coarse image with multiple objects related to the task. The authors show that Contextual RNN initialised with contextual information from input sequence improves the performance for an associative retrieval task compared to the baseline with zero initialisation.

Next, we consider recent literature in various other domains where sensory data is used as a context for RNN models.  In \citep{Batbaatar2018},  a LSTM model is used to predict appliances energy. Energy usage of home appliances is highly dependent on weather conditions. In this work, two different datasets are combined with the aim of introducing contextual features to the LSTM model. The contextual features are obtained from the first dataset that consists of house temperature and humidity measures for a period of 4.5 months. The second dataset provides the measurements of individual household electric power consumption over a period of 47 months. The work focused on identifying the best predictors for the prediction task and did not attempt to adapt the model for incorporating contextual information. Lee and Hauskrecht \citep{Min2019}, proposed a LSTM-based model for predicting event occurrence in clinical event time series with events such as medication orders, lab tests and their results, or physiological signals. The proposed model considers the usual abstracted information from the past through the hidden states, and also recent event occurrence information. The recent event at the current time step is represented as a multi-hot vector which is linearly transformed and added to the LSTM model. The output from the model depends on both the hidden state and the current event information. The authors demonstrated through experimental results with MIMIC-III clinical event data \cite{Johnson2016} that by combining recent event information and the summarised past events through LSTM hidden states improves the prediction over models that do not incorporate the former information.

In \citep{Belhadi2020}, the task of long-term traffic flow (traffic flow over long interval) prediction is performed using a RNN that considers inputs from multiple data sources. The RNN uses, flow information, temporal information, and contextual information.  The type of the day represents the contextual feature. For example, weekend or regular and event or no-event day (celebration days). Binary values are used for representation, 0 is used for weekend and event day and 1 is used for regular and non-event days. In this work also no emphasis is given on contextual learning. It simply extends the input feature vector to accommodate the contextual features. Ma et al. \citep{Ma2021} also proposed a contextual convolutional RNN for long-term (daily) traffic flow forecasting. The traffic flow data is represented in a form such that it is similar to a sequence of square images which is given as input to the CNN. The CNN is used to extract the inter-and intra-day traffic pattern. The output of CNN is given as input to a LSTM network. Here, four contextual features are used, day of the week, season, weather, and holiday. The contextual features are represented as one-hot encoding.  Finally, contextual features and the output from the LSTM are concatenated and given as input to a linear dense layer for the prediction of traffic flow. The performance of the model was evaluated using real data of Seattle city. The results show that the proposed model provides better prediction particularly during high demand periods. In \citep{QU2021}, a Feature injected RNNs (Fi-RNN) is presented to predict short-term traffic speed. The model consists of an RNN, an autoencoder, and a feedforward neural network. The historical traffic flow data is given as input to the RNN (LSTM) to learn the sequential pattern in terms of sequential features. The contextual data is used by a sparse autoencoder to obtain an expanded set of contextual features. The expanded contextual features are restricted to [0,1] using a softmax function and this value is used to weight the features from LSTM (the hidden states of LSTM). Finally, these weighted feature vectors are used as input to the feed-forward neural network to learn traffic patterns and predict the speed. The experimental results from two real datasets show that Fi-RNN that incorporates contextual features achieves improved accuracy. 

To summarize, all the RNN models that are discussed in this section obtained improved results by adding additional contextual information to the model. In these approaches, the context information has been added primarily in three ways to the RNN model. Let us denote the primary feature vector at time $t$ as $\mathbf{x}_t \in \mathbb{R}^{(n_x \times 1)}$ and contextual feature vector as $\mathbf{z}_t \in \mathbb{R}^{(n_z \times 1)}$. Firstly, most of the approaches, integrate context with the primary input to get an extended input feature vector. The extended vector can be represented as $\mathbf{x}\;' = [\mathbf{x}_t,\mathbf{z}_t]$ where $\mathbf{x}\;'_t \in \mathbb{R}^{(n_x+n_z) \times 1}$. Sometimes, the new input vector is obtained multiplicatively as $\mathbf{x}\;' = [\mathbf{x}_t \odot \mathbf{z}_t]$. Note that in this case, either the context feature or both primary and context features are transformed such that they are of the same dimension. In the second manner, context is introduced additively and provided as a separate input to the hidden or output layer or both. The hidden state is computed as $f(\mathbf{W}\mathbf{x}_t + \mathbf{U}\mathbf{h}_{t-1} + \mathbf{V}\mathbf{z}_t)$ and when context is added to the output, it is computed as $g(\mathbf{U}\mathbf{h}_t + \mathbf{V}\mathbf{z}_t)$.  Here, $\mathbf{W}, \mathbf{U}, \mathbf{V}$ are the associated weights and $f, g$ are the activation functions. Finally, the context is integrated multiplicatively in the hidden unit as as $f(\mathbf{W}\mathbf{x}_t \odot  \mathbf{V}\mathbf{z}_t + \mathbf{U}\mathbf{h}_{t-1} \odot  \mathbf{V}\mathbf{z}_t)$ and the output as $g(\mathbf{U}\mathbf{h}_t \odot \mathbf{V}\mathbf{z}_t)$.

As mentioned briefly in section \ref{sec:introduction}, in contrast to the above approaches, in this paper we integrate context such that the weights associated with the primary input vector is dependent on the context. Accordingly, the hidden state is given as, $f(\mathbf{W}(\mathbf{z}_t)\mathbf{x}_t + \mathbf{U}\mathbf{h}_{t-1})$ and the output unit computation remains same as RNN. This will be discussed in the next section in more detail.

\section{Proposed Approach}
\label{sec:proposed_approach}
In this section, we describe the proposed Context integrated Recurrent Neural Network (CiRNN) model which uses GRUs as the basic unit of the network.  We first give the notations, and then briefly discuss the GRU to relate it to CiRNN. Finally, we present the details of CiRNN.

\subsection{Notations}
\label{subsec:notations}

The following notations are used to describe CiRNN.
\begin{equation}
    \begin{split}
        & n_x : \text{primary input dimension} \\
        & n_h : \text{number of hidden units} \\
        & n_y: \text{output dimension} \\ 
        & n_z : \text{context input dimension} \\
        & T_x : \text{input sequence length} \\ 
        & T_y: \text{output sequence length} \\
        & (\text{for many-to-one RNN architecture $T_y = 1$}) \\
        & \mathbf{x}_t \in \Re^{(n_x \times 1)} : \text{input vector at time step $t$} \\
        & \mathbf{z}_t \in \Re^{(n_z \times 1)} : \text{context vector at time step $t$} \\
        & \mathbf{y}_t \in \Re^{(n_y \times 1)}: \text{output at time step $t$} \\
        & \hat{\mathbf{y}}_t \in \Re^{(n_y \times 1)}: \text{estimated output at time step $t$} \\
        & \mathbf{h}_t \in \Re^{(n_h \times 1)}: \text{hidden unit activation  at time step $t$} \\
        & \tilde{\mathbf{h}}_t \in \Re^{(n_h \times 1)}: \text{candidate activation  at time step $t$} \\
        & \mathbf{s}_t \in \Re^{(n_h \times 1)}: \text{set/update gate at time step $t$} \\
        & \mathbf{r}_t \in \Re^{(n_h \times 1)}: \text{reset gate at time step $t$} \\
        & \mathbf{W}^{s} \in \Re^{(n_h \times n_x)}: \text{weights (set gate to input)} \\
        & \mathbf{W}^{h} \in \Re^{(n_h \times n_x)}: \text{weights (hidden to input)} \\
        & \mathbf{W}^{r} \in \Re^{(n_h \times n_x)}: \text{weights (reset gate to input)} \\ 
        & \mathbf{U}^{s} \in \Re^{(n_h \times n_h)}: \text{weights (set gate  to hidden)} \\
        & \mathbf{U}^{h} \in \Re^{(n_h \times n_h)}: \text{weights (hidden to hidden)} \\
        & \mathbf{U}^{r} \in \Re^{(n_h \times n_h)}: \text{weights (reset gate to hidden)} \\
        & \mathbf{V} \in \Re^{(n_y \times n_h)}: \text{weights (output to hidden)} \\
        & \mathbf{b}_y \in \Re^{(n_y \times 1)} : \text{bias of output unit} \\
    \end{split}    
\end{equation}

\subsection{Gated Recurrent Units}
\label{subsec:GRU}

A GRU \citep{cho-etal-2014} is a unit or cell, which is used as a hidden unit in RNNs. It can adaptively remember and forget the information in the network during the learning phase. The primary motivation of GRU is same as LSTM, which is to model long-term sequences while avoiding the vanishing gradient problem of RNN. However, compared to LSTM, GRU does this with less number of parameters and less number of operations. GRU comprises of two gates, reset and update gate that controls the flow of information. For time step t, the GRU computes the output $\hat{\mathbf{y}}_t$ using input $\mathbf{x}_t$ and the hidden state  as follows:
\begin{equation}
    \begin{split}
       & \hat{\mathbf{y}_t}  = f(\mathbf{V} \mathbf{h}_t + \mathbf{b}_y) \\
       &  \mathbf{h}_t  = \mathbf{s}_t \odot \mathbf{h}_{t-1} + (1-\mathbf{s}_t) \odot \tilde {\mathbf{h}}_t\\
       & \tilde {\mathbf{h}}_t  = tanh(\mathbf{W}^h \mathbf{x}_t +  \mathbf{U}^h (\mathbf{r}_t \odot \mathbf{h}_{t-1})) \\
       & \mathbf{s}_t  = \sigma (\mathbf{W}^s \mathbf{x}_t + 
         \mathbf{U}^s \mathbf{h}_{t-1}) \\
       & \mathbf{r}_t  = \sigma(\mathbf{W}^r \mathbf{x}_t + \mathbf{U}^r \mathbf{h}_{t-1}) 
    \end{split}
\end{equation}

The reset gate functions very similar to the forget gate of the LSTM cell. It is a linear combination of input at time step $t$ and hidden state at previous time step $t-1$. The reset gate resets the hidden state with current input and ignores the information from previous hidden state when its value is close to 0. It controls how much information from the previous hidden state will be removed. On the other hand, the update gate controls how much information from the previous hidden state will be remembered. 

\subsection{Context Integrated RNN}
\label{subsec:CiRNN}
\begin{figure*}[tb]
\centering
\includegraphics[width=6.0in]{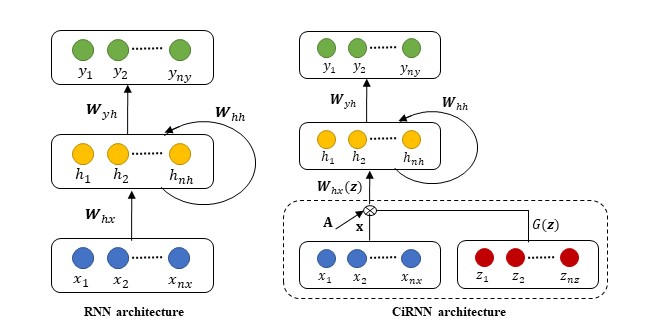}
\caption{Basic architecture of RNN vs. CiRNN. The connection weights between various layers is represented by $\mathbf{W}$. In CiRNN, the input to hidden layer connection weights ($\mathbf{W}_{hx}$) are dependent on the context $\mathbf{z}$ \citep{Baruah2023}}.
\label{fig:basic_arch}
\end{figure*}

Context Integrated RNN is a RNN with GRU as hidden units. For clarity of presentation, in Figure \ref{fig:basic_arch}, the basic architecture of a RNN is compared with CiRNN. The main difference between the two architectures is that in CiRNN, the input to hidden unit weights are dependent on the context variables. When each of the hidden units $\mathbf{h}_1, \mathbf{h}_2, \dots \mathbf{h}_{nh}$ in Figure \ref{fig:basic_arch} is a GRU then the parameters $\mathbf{W}^s, \mathbf{W}^r, \mathbf{W}^h$ are dependent on the context variables $\mathbf{z}$. 

The output computation in a single unit of CiRNN is same as in GRU. However, the computation of candidate hidden state, set/update gate, and reset gate values are computed differently as it involves weights that are dependent on context $\mathbf{z}_t$ as shown below:
\begin{equation}
\label{eqn:all_eqn}
    \begin{split}
       & \hat{\mathbf{y}_t}  = f(\mathbf{V} \mathbf{h}_t + \mathbf{b}_y) \\
       &  \mathbf{h}_t  = \mathbf{s}_t \odot \mathbf{h}_{t-1} + (1-\mathbf{s}_t) \odot \tilde {\mathbf{h}}_t\\
       & \tilde {\mathbf{h}}_t  = tanh(\mathbf{W}^h(\mathbf{z}_t) \mathbf{x}_t +  \mathbf{U}^h (\mathbf{r}_t \odot \mathbf{h}_{t-1})) \\
       & \mathbf{s}_t  = \sigma (\mathbf{W}^s(\mathbf{z}_t) \mathbf{x}_t + 
         \mathbf{U}^s \mathbf{h}_{t-1}) \\
       & \mathbf{r}_t  = \sigma(\mathbf{W}^r (\mathbf{z}_t)\mathbf{x}_t + \mathbf{U}^r \mathbf{h}_{t-1}) 
    \end{split}
\end{equation}
\noindent
In equation \ref{eqn:all_eqn}, the weights associated with the input $(\mathbf{x}_t)$ are dependent on the vector of contextual variables $(\mathbf{z}_t)$. Let us consider one of the such parameters, $\mathbf{W}^s$. Note that $\mathbf{W}^h(\mathbf{z}_t)$ and $\mathbf{W}^r(\mathbf{z}_t)$ can be expressed in similar way.

The matrix $\mathbf{W}^s(\mathbf{z}_t)$ is of dimension $n_h \times n_x$ and each of the components can be given as:
\begin{equation}
  \mathbf{W}^s(\mathbf{z}_t) =   \begin{bmatrix}
                                     w_{11}^s(\mathbf{z}_t) & w_{12}^s(\mathbf{z}_t) & \cdots & w_{1n_x}^s(\mathbf{z}_t)\\
                                     w_{21}^s(\mathbf{z}_t) & w_{22}^s(\mathbf{z}_t) & \cdots & w_{2n_x}^s(\mathbf{z}_t)\\
                                     \vdots & \vdots & \cdots & \vdots \\
                                     w_{n_h1}^s(\mathbf{z}_t) & w_{n_h2}^s(\mathbf{z}_t) & \cdots & w_{n_hn_x}^s(\mathbf{z}_t)\\
                                     
                                \end{bmatrix}
\end{equation}

\noindent
where each element of the matrix can be expressed as:

\begin{equation}
\label{eqn:weight_elt}
\begin{split}
    &w_{ki}^s = \mathbf{A}_{ki}^s \mathbf{G}(\mathbf{z}_t), \quad k = 1,..,n_h, \quad i = 1,..,n_x \\
    &\mathbf{A}_{ki}^s = [a_{ki1}^s, a_{ki2}^s, \cdots, a_{kim}^s]
\end{split}
\end{equation}

In (\ref{eqn:weight_elt}), $\mathbf{G}(\mathbf{z}_t) = [g_1(\mathbf{z}_t), g_2(\mathbf{z}_t),...,g_m(\mathbf{z}_t)]^T$ is a vector of basis functions that can be chosen at the time of design and $\mathbf{A}_{ki}^s$ is a vector of coefficients that specify the dependence of weights on context variables \citep{Ciskowski2004}. We can define a matrix $\mathbf{A}^s$ where each row $\mathbf{A}^s_k$ can be formed by concatenating coefficient vectors $\mathbf{A}_{ki}^s$ as shown below. Therefore, $\mathbf{A}^s$ is of dimension $(n_h \times n_xm)$. 
\begin{equation}
    \mathbf{A}_k^s = [\mathbf{A}_{k1}^s, \mathbf{A}_{k2}^s, \cdots, \mathbf{A}_{kn_x}^s], \quad k = 1,2,..., n_h
\end{equation}

Using $\mathbf{A}^s$ and similarly $\mathbf{A}^h$ and $\mathbf{A}^r$, the candidate hidden state $\tilde{\mathbf{h}}_t$, the update (set) gate $\mathbf{s}_t$, and reset gate $\mathbf{r}_t$ in equation (\ref{eqn:all_eqn}) can be expressed as:

\begin{equation}
\label{eqn:all_eqn_new}
    \begin{split}
       & \tilde{\mathbf{h}_t}  = tanh[\mathbf{A}^h (\mathbf{x}_t \otimes \mathbf{G}(\mathbf{z}_t)) + \mathbf{U}^h (\mathbf{r}_t \odot \mathbf{h}_{t-1})] \\
       & \mathbf{s}_t  = \sigma [\mathbf{A}^s (\mathbf{x}_t \otimes \mathbf{G}(\mathbf{z}_t)) +  \mathbf{U}^s \mathbf{h}_{t-1}] \\
       & \mathbf{r}_t  = \sigma[\mathbf{A}^r (\mathbf{x}_t \otimes \mathbf{G}(\mathbf{z}_t)) + \mathbf{U}^r \mathbf{h}_{t-1}] 
    \end{split}
\end{equation}

In (\ref{eqn:all_eqn_new}), $ \otimes $ denotes Kronecker product of matrices. Learning of each of the weights $w_{ki}^s$ will require learning of the vector of coefficients $\mathbf{A}_{ki}^s$ with $m$ elements. The parameter learning process in CiRNN is similar to RNNs which requires defining a loss function and optimization of parameters using any suitable optimization algorithm such as stochastic gradient descent (SGD) or Adam optimization algorithm \citep{Kingma2014} with  Back Propagation Through Time (BPTT or Truncated BPTT). (The gradient calculation of L2 Loss function is provided in the Appendix A.)


\section{Experiments and Results} 
In this section, we first describe the evaluation task and the dataset, and then discuss the experiments and the results achieved with the proposed model. The results of CiRNN are compared with baseline models and also with the state-of-art methods from the existing literature. 

\subsection{Predictive Maintenance}
For evaluating the efficacy of CiRNN, we considered the task of predictive maintenance. Predictive maintenance (PdM) also referred to as Prognostic and Health Management (PHM)  has become more relevant with the arrival of Industry 4.0 within the context of smart manufacturing and industrial big data. Through PdM, various industries aim to achieve near-zero failures, downtime, accidents, and unscheduled maintenance thereby saving considerable amount of cost and also improve operational safety \citep{Zhang2019}.

PdM primarily comprises of diagnostics, and prognostics mechanisms. The former deals with fault detection, isolation and identification and the latter attends to degradation monitoring and failure prediction. Compared to diagnostics, prognostics is more efficient to achieve zero-downtime performance \citep{SHIN2015}. Prediction of Remaining Useful Life (RUL) is one of the crucial tasks in prognostics. RUL, also called remaining service life, residual life or remnant life, refers to the time left before observing a failure given the current machine age and condition, and the past operation profile \citep{JARDINE2006}. A reliable estimation of RUL enables the stakeholders to attain the benefits of PdM. 

In recent years, industrial cyber-physical system has emerged as one of the key technologies for reliable data acquisition in real-time. Thanks to such systems that made huge amount of data available which in turn led to a growing interest among researchers to apply machine learning, in particular deep learning, approaches to industrial applications. A review of deep neural networks applied to PHM is available in \citep{REZAEIANJOUYBARI2020} and \citep{Zhang2019}.  In \citep{WANG2020} the state-of-the-art deep learning approaches for RUL prediction is discussed. Here, we briefly provide an overview of some of the recent approaches for prediction of RUL that employ C-MAPSS dataset as our work is evaluated with the same dataset. More details on the dataset is provided in the next section. In \citep{Heimes2008}, one of the early works that used a basic RNN for estimation of RUL of turbofan engines is presented. It is based on a RNN trained with back-propagation through time using Extended Kalman Filter. This work was placed second in the IEEE 2008 Prognostics and Health Management conference challenge problem \citep{Jia2018}.

In \citep{Yuan2016} and \citep{WU2018}, a vanilla LSTM is used for prediction of RUL. They showed that LSTM could perform better compared to basic RNNs, GRUs and Adaboost-LSTM. In \citep{Zheng2017}, deep LSTM with a feedforward newtwork is used to estimate the RUL. The six operating conditions of the engine are encoded as one hot vector and are used while training the model. The results showed that deep LSTM model with feedforwarded network performed better compared to Multi-layer Perceptron, Support Vector Regression, Relevance Vector Regression, and Deep Convolutional Neural Network (CNN). Li et al. \citep{LI2018} applied 1D convolutions in sequence without pooling operations for useful feature extraction in RUL prediction. The model achieved good results without incurring high training time compared to recurrent models. Listou Ellefsen et al., \citep{LISTOUELLEFSEN2019}, used a Restricted Boltzmann Machine (RBM) to investigate the effect of unsupervised pre-training in RUL predictions. Initially, RBM is used to get the initial weights from unlabeled data and later fine tuning is performed in an LSTM network with labeled data. Further, a Genetic Algorithm (GA) approach is applied for tuning the hyper-parameters during the training. The model proposed in \citep{ZHAO2019}, emphasizes on trend features. The trend features are extracted using Complete Ensemble Empirical Mode Decomposition (CEEMD) approach, followed by reconstruction procedure. These features are used to train a LSTM for RUL prediction. A global attention mechanism is used with LSTM network in \citep{Costa2019}. It learns the input to RUL mapping directly from the raw sensor data and do not require feature engineering or unsupervised pre-training. Further, the attention mechanism allows visualising the learned attention weights at each step of prediction of RUL.

The existing approaches discussed here consider the C-MAPSS dataset. The dataset consists of three operational setting parameters that provide information about different flight operating conditions which is contextual information. The existing approaches do not explicitly utilize this information during the learning process. They either excluded this information and used only the sensor data during training \citep{ZHAO2019} or included them as primary features \citep{Yuan2016}, \citep{Costa2019}.

\subsection{Dataset Description}
NASA Turbofan Engine Degradation Simulation Data Set \citep{Saxena2008} is a widely used benchmark for RUL prediction. The dataset is generated using Commercial Modular Aero-Propulsion System Simulation (C-MAPSS) tool. It consists of four distinct datasets (FD001, FD002, FD003, and FD004) that contain information from 21 sensors (such as Total temperature at fan inlet, Total temperature at Low Pressure Compressor outlet), 3 operational settings (flight altitude, Mach number, and throttle resolver angle), engine identification number, and cycles of each engine. The degradation in engine performance is captured by the sensor data. The engine performance is also substantially influenced by the three operational settings.  FD001 has one operating condition and one failure mode. FD002 has six operating conditions and one failure mode. FD003 has one operating condition and two failure modes and FD004 has six operating conditions and two failure modes. The details are provided in Table \ref{tab:dataset}. For our approach, both FD002 and FD004 are suitable as the six operating conditions provide the contextual information. However, for experiments we have also considered FD001 and FD003 to evaluate the performance where only one context is available (single operating condition).

The engines start with a different degree of initial wear and manufacturing variation which is not known to the user and are not considered as a fault condition. Each of the datasets consist of separate training and test sets. In the training set, the sensor data is captured until the system fails whereas in the test set it is captured up to a certain time prior to the failure. The test sets also have the true Remaining Useful Life (RUL) values.
 
The objective is to predict the number of remaining operational cycles before failure as provided in the test set.

\begin{table}
\centering
\renewcommand{\arraystretch}{1.3}
\caption{Specification of C-MAPSS datsets}
\label{tab:dataset}
\begin{tabular}{l l l l l }
\hline
\bfseries Specification & \bfseries FD001  & \bfseries FD002
& \bfseries FD003 & \bfseries FD004\\
\hline
Engine units (Training) & 100 & 260 & 100 & 249 \\
Engine units (Testing) & 100 & 259 & 100 & 248 \\
Operating Conditions & 1 & 6 & 1 & 6\\
Fault Modes & 1 & 1 & 2 & 2\\
\hline
\end{tabular}
\end{table}

\subsection{Data Preprocessing}

For each of the dataset, univariate and bivariate analysis of data from the 21 sensors is performed. Each of the sensor data series is analyzed to see the trend with respect to the time cycles. Further, for each dataset, sensors are selected based on the scatter plot observations and correlation analysis. The selected sensors and operational settings is provided in Table \ref{tab:sensors}. The sensor values are used as primary features and the operational settings are used as contextual features while learning the models. 

\begin{table}
\centering
\renewcommand{\arraystretch}{1.3}
\caption{Selected sensors and operational settings}
\label{tab:sensors}
\centering
\begin{tabular}{p{1.5cm} p{5cm} p{5cm}}
\hline
\bfseries Dataset & \bfseries Sensors  & \bfseries Operational Settings\\
\hline
FD001 & s2, s3, s4, s7, s8, s9, s11, s12, s13, s15, s17, s20, s21 & OS1, OS2 \\
FD002 & s1, s2, s8, s13, s14, s19 & OS1, OS2, OS3 \\
FD003 & s2, s3, s4, s7, s8, s9, s11, s15, s17, s20, s21 1 & OS1,OS2\\
FD004 & s2, s8, s14, s16 & OS1, OS2, OS3 \\
\hline
\multicolumn{3}{p{11.5cm}}{OS1- Flight altitude, OS2- Mach number, OS3- Throttle resolver} \\
\end{tabular}
\end{table}

\begin{figure}
\centering
\includegraphics[width=3.3in, height = 2.5in ]{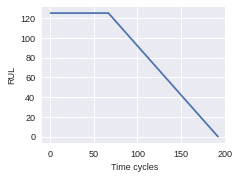}
\caption{RUL of engine unit 1 (FD002 training data)}
\label{fig:RUL}
\end{figure}

The RUL for training is required to be identified as the true RUL values are not provided in the training data. We adopted a piece-wise linear degradation model to get the RUL values for the training data as given in \citep{LISTOUELLEFSEN2019} and \citep{Costa2019} . The degradation model assumes that after an initial period with constant RUL values (initially when the engine operates in normal conditions) \citep{Heimes2008} the RUL decrease linearly as there is progress in the number of time cycles. For our experiments, the initial constant RUL is considered to be 125 cycles so that results can be compared with existing works. Figure \ref{fig:RUL} shows the RUL for engine unit 2 of FD002 training data. 

For FD001 and FD003 (with only one operational condition), we performed min-max normalisation on both primary and contextual inputs, and also on the target. For FD002 and FD004 (with six operational conditions) we performed contextual normalisation \citep{Turney1996} in addition to min-max normalisation (excluding the target). For contextual normalisation, the data is clustered according to 6 operational regimes using k-means clustering and then the data is normalised using the cluster statistics (mean and range). Finally, data smoothing is performed using moving averages with window size of 3.

\subsection{Performance Metrics}
Two metrics are used to measure the performance of the proposed model. The first metric is the RMSE,  and the second metric is a score that is computed from an asymmetric scoring function as proposed by Saxena et al. \citep{Saxena2008}, which is given as,

\begin{equation}
\label{eqn:score}
  \setlength{\arraycolsep}{0pt}
    s = \left\{ \begin{array}{l l}
    \sum_{i=1}^{n} e^{- \frac{D_i}{a_1}} -1, \ \text{if} \ D_i < 0 \\
     \sum_{i=1}^{n} e^{\frac{D_i}{a_2}} -1, \ \text{if} \ D_i \geq  0
  \end{array} \right.
 \end{equation}

where $a_1 = 10$, $a_2 = 13$, and $D_i = \hat{RUL}_i - RUL_i$ is the difference between predicted RUL and actual RUL values, $n$ is the number of samples in the test data. The scoring metric penalizes the late predictions (positive errors) more compared to early predictions (negative errors). In either case, the penalty increases exponentially with error.

\subsection{Training and Validation}

To train the models, the training and validation data is prepared in the following way. From each of the dataset, each engine unit is considered and last $k$ samples are selected for validation and remaining are used for training. It is to be noted that $k$ is required to be multiple of sequence length used while training the models. As a result, the validation set consists of samples from each engine unit as in the test set. Table \ref{tab:param} shows various hyperparameters that are used for tuning the models. The number of layers is fixed to one, due to computational overhead of custom built CiRNN code. For the contextual inputs, polynomial basis functions of degree 2 are used. 

\begin{table}
\centering
\renewcommand{\arraystretch}{1.3}
\caption{Hyperparameter values and optimizer used for tuning}
\label{tab:param}
\centering
\begin{tabular}{l l}
\hline
\bfseries Hyperparameter & \bfseries Value \\
\hline
Number of hidden units & \{10, 15, 20, 25, 30 \} \\
Sequence length & \{10, 15, 20\} \\
Batch size & \{64, 128, 256 \} \\
Learning rate & loguniform(1e-5, 1e-3)\\
\hline
Optimizer & SGD, Adam, RMSProp \\
\hline
\end{tabular}
\end{table}

We considered GRU-RNN as a baseline where the RNN consits of GRU but without contextual weighting. Both baseline and CiRRN are trained with various configurations. For example, GRU-RNN with contextual normalisation and with context features where context features are treated as primary features. The later is also referred to as contextual expansion \citep{Turney1996}. Table \ref{tab:models} shows various models and Table \ref{tab:models_hyperparams} presents the best hyperparameter values for each model that are achieved after tuning them.  In the next subsection, we discuss the results obtained with these models.

\begin{table}[tb]
\centering
\caption{Model configurations}
\label{tab:models}

\begin{tabular}{p{2.5cm} l l p{2.8cm} p{1.8cm}}
\hline
Dataset & Model & Model Acronym & Contextual Normalisation & Context Features\\
\hline
FD001  & CiRNN & CiRNN\_D1 & No  &  Yes \\
 (1 operational condition) & GRU & GRU\_D1\_CxF & No & Yes \\
 & GRU & GRU\_D1 & No & No \\
 \hline
 FD002  & CiRNN & CiRNN\_D2 & Yes & Yes \\
 (6 operational conditions) & CiRNN & CiRNN\_D2\_CxF & No & Yes \\
 & GRU & GRU\_D2 & No & No \\
 & GRU & GRU\_D2\_CxF & No & Yes \\
 & GRU & GRU\_D2\_CxN & Yes & No \\
 & GRU & GRU\_D2\_CxN\_CxF & Yes & Yes \\
 \hline
 FD003  & CiRNN & CiRNN\_D3 & No  &  Yes \\
 (1 operational condition) & GRU & GRU\_D3\_CxF & No & Yes \\
 & GRU & GRU\_D3 & No & No \\
 \hline
  FD004  & CiRNN & CiRNN\_D4 & Yes & Yes \\
 (6 operational conditions) & CiRNN & CiRNN\_D4\_CxF & No & Yes \\
 & GRU & GRU\_D4 & No & No  \\
 & GRU & GRU\_D4\_CxF & No & Yes \\
 & GRU & GRU\_D4\_CxN & Yes & No \\
 & GRU & GRU\_D4\_CxN\_CxF & Yes & Yes \\
 \hline
 
\end{tabular}
\end{table}

\begin{table*}[tb]
\centering
\caption{Model configurations and Hyperparameters}
\label{tab:models_hyperparams}
\centering
\begin{tabular}{l p{4.5cm} l}
\hline
Model Acronym & Hyperparameter values & Optimizer\\
\hline
CiRNN\_D1 & $15, 20, 64, 1\times10^{-2}$ & Adam\\
 GRU\_D1\_CxF & $25, 20, 128, 5\times10^{-3}$ & Adam\\
 GRU\_D1 & $10, 15, 64, 9\times10^{-3}$ & Adam\\
 \hline
 CiRNN\_D2 & $20, 15, 64, 5\times10^{-3},$  & RMSProp \\
 CiRNN\_D2\_CxF & $15, 20,128, 5\times10^{-3}$ & RMSProp \\
 GRU\_D2 & $25, 15, 128, 1\times10^{-2}$ & Adam\\
 GRU\_D2\_CxF & $30, 15, 128, 5\times10^{-3}$ & Adam\\
 GRU\_D2\_CxN & $20, 10, 64, 8\times10^{-3}$ & RMSProp \\
 GRU\_D2\_CxN\_CxF & $15, 10, 64, 9\times10^{-3}$ & RMSProp \\
 \hline
 CiRNN\_D3 & $30, 10, 64, 8\times10^{-3}$ & RMSProp \\
 GRU\_D3\_CxF & $15, 10, 64, 2\times10^{-3}$ & RMSProp\\
 GRU\_D3 & $20, 10, 64, 5\times10^{-3}$ & RMSProp \\
 \hline
  CiRNN\_D4 & $15, 10, 128, 8\times10^{-3}$ & RMSProp \\
 CiRNN\_D4\_CxF & $25, 20, 128, 9\times10^{-3}$ & Adam\\
 GRU\_D4 & $25, 20, 128, 8\times10^{-3}$ & Adam \\
 GRU\_D4\_CxF & $25, 15, 256, 5\times10^{-3}$ & Adam \\
 GRU\_D4\_CxN & $20, 15, 256, 1\times10^{-2}$ & RMSProp\\
 GRU\_D4\_CxN\_CxF & $15, 15, 256, 9\times10^{-3}$ & Adam\\
 \hline
 \multicolumn{3}{p{11.5cm}}{Hyperparameter values: hidden units, sequence length, batch size, learning rate} \\
\end{tabular}
\end{table*}

\subsection{Results}
\begin{figure*}[t!]
    \centering
     \begin{subfigure}[t]{0.8\textwidth}
        \centering
        \includegraphics[width = 5in, height=1.8in]{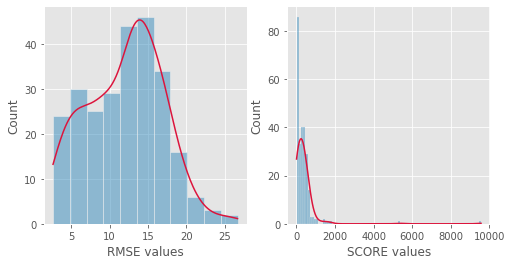}
        \caption{ }
    \end{subfigure}\\%
    
    \begin{subfigure}[t]{0.8\textwidth}
        \centering
        \includegraphics[width = 5in, height=1.8in]{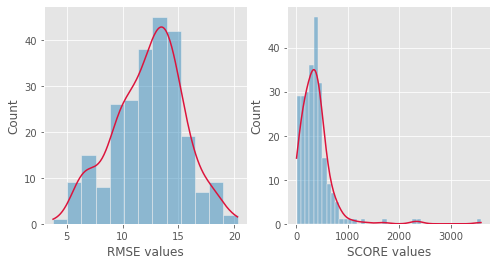}
        \caption{ }
    \end{subfigure}
    
    \caption{Distribution of RMSE and Score values obtained from (a) CiRNN\_D2 with FD002 test set (b) CiRNN\_D4 with FD004 test set}
    
\label{fig:dist}
\end{figure*}

Table \ref{tab:results1} summarises the results obtained from CiRNN and the baseline models with the test dataset. The testing is performed for each engine unit separately and the average RMSE and average score with standard deviation is reported. For FD001 dataset, which has only one operational condition, RNN with GRU model where context features are added and treated as primary features (GRU\_D1\_CxF) performed better than other two models.  The percentage increase in RMSE for CiRNN\_D1 in comparison to GRU\_D1\_CxF is around 7\%.  For all the remaining datasets (FD002, FD003, FD004) CiRNN performed better or at par with the baseline models. It can be observed that contextual normalisation along with contextual weighting through CiRNN, particularly in datasets FD002 and FD004, have noticeably improved the performance. Further, it is to be noted that the distribution of the score metric is skewed. Here, we have provided distribution of RMSE and scores for FD002 and FD004 with CiRNN in Figure \ref{fig:dist}. It is apparent from the figures that the count of engines having high scores (more than 1000) is very low. 

\begin{table*}[t!]
\renewcommand{\arraystretch}{1.3}
\caption{Comparison of CiRNN performance with baseline models}
\label{tab:results1}
\centering
\begin{tabular}{l l l l}
\hline
Dataset & Model & RMSE & Score $(s)$ \\
 & & (mean, std) & (mean, std) \\
\hline
 FD001 & CiRNN\_D1 & 14.53, 7.07 & 745.21, 2588.64 \\
 & GRU\_D1 & 14.52, 7.64 & 658.55, 1490.477 \\
 & \textbf{GRU\_D1\_CxF} & \textbf{13.57, 7.19} & \textbf{450.69, 733.00} \\
  \hline
 FD002  & \textbf{CiRNN\_D2} & \textbf{11.97, 4.94} & \textbf{363.03, 710.70}\\
 & CiRNN\_D2\_CxF & 25.74, 10.11 & 4123.71, 8035.64\\
 & GRU\_D2 & 26.57, 7.78 & 3156.17, 5158.81\\
 & GRU\_D2\_CxF & 25.83, 10.04 & 4122.89, 7878.20 \\
 & GRU\_D2\_CxN & 12.49, 4.91 &  417.26, 841.84 \\
 & GRU\_D2\_CxN\_CxF & 12.61, 5.09  & 470.85, 1006.85\\
 \hline
 FD003  & \textbf{CiRNN\_D3} & \textbf{12.87, 10.77} & \textbf{949.39, 2437.71} \\
 & GRU\_D3 &13.23, 11.26 & 1633.70, 7559.84\\
 & \textbf{GRU\_D3\_CxF} & \textbf{12.90, 3.39} & \textbf{902.63, 2339.36}\\
 \hline
  FD004  & \textbf{CiRNN\_D4} & \textbf{12.35, 3.06}  & \textbf{377.94, 353.19}\\
 & CiRNN\_D4\_CxF & 23.11, 8.91 & 3730.99, 15989.09\\
 & GRU\_D4 & 22.87, 10.92 & 4604.87, 13398.62\\
 & GRU\_D4\_CxF & 21.92, 9.55& 3539.59, 12060.03\\
 & GRU\_D4\_CxN & 16.91, 4.70 & 797.56, 1360.32\\
 & GRU\_D4\_CxN\_CxF & 17.75, 4.29& 788.20, 678.51\\
 \hline
 \end{tabular}
\end{table*}

\begin{figure}[tb]
\centering
\includegraphics[width=3.5in, height = 2.5in ]{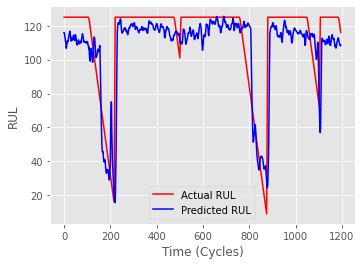}
\caption{Actual versus Predicted RUL with CiRNN using FD004 test data}
\label{fig:act_vs_pred}
\end{figure}

\begin{figure*}[t!]
    \centering
     \begin{subfigure}{0.5\textwidth}
        \centering
        \includegraphics[height=2in]
        {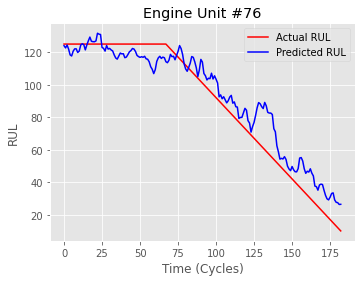}
        \caption{ }
    \end{subfigure}%
    \begin{subfigure}{0.46\textwidth}
        \centering
        \includegraphics[height=2in]
        {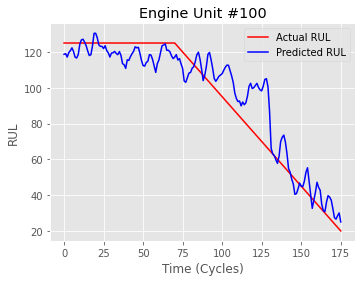}
        \caption{ }
    \end{subfigure}
    \begin{subfigure}{0.5\textwidth}
        \centering
        \includegraphics[height=2in]{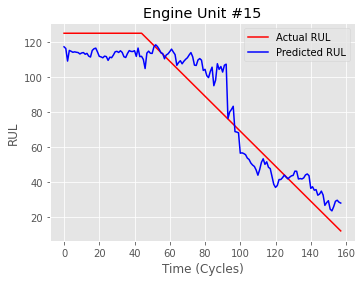}
        \caption{ }
    \end{subfigure}
    \begin{subfigure}{0.46\textwidth}
        \centering
        \includegraphics[height=2in]{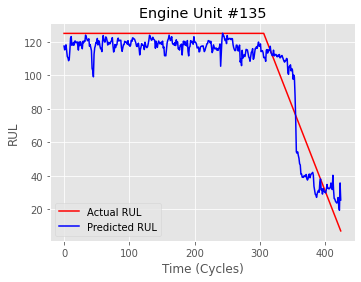}
        \caption{ }
    \end{subfigure}
    \caption{Actual versus predicted RUL for selected engines (a) and (b) CiRNN with FD001 test data, (c) and (d) CiRNN with FD004 test data}
    
\label{fig:engine_rul}
\end{figure*}

Figure \ref{fig:act_vs_pred} shows the predicted RUL values versus actual RUL values for the first 1200 samples of the FD004 test data and in Figure \ref{fig:engine_rul} the predicted and actual RUL values of two randomly selected engine units from FD001 and FD004 test data is shown.  It can be noticed from Figure \ref{fig:act_vs_pred} that for longer running times, the model prediction is better compared to shorter running times in FD004. The late predictions in case of shorter running times also contribute to higher score from scoring function. If we observe the predictions of engine units closely (Figure \ref{fig:engine_rul}), late predictions are more in FD001 compared to FD004. This is also observed in Table \ref{tab:results1}, where CiRNN predictions have much higher scores in datasets with one operating condition due to late predictions. 

A comparison of results achieved from CiRNN and results from state-of-the art deep learning models applied to the same dataset is presented in Table \ref{tab:results2} (FD001 and FD003) and Table \ref{tab:results3} (FD002 and FD004). The models that are selected for comparison are sequential models based on LSTM \citep{Zheng2017}, \citep{LISTOUELLEFSEN2019}, \citep{Costa2019}, \citep{ZHAO2019} and additionally CNN \citep{LI2018} is considered. It is worth mentioning here that there are several works reported in the literature that use C-MAPSS dataset, however, most of the works are based on only FD001 dataset and are not tested on all the four datasets. As it can be seen from the Table \ref{tab:results3}, CiRNN performed better in both the datasets FD002 and FD004 where contextual information in terms of operational settings is available. Our model resulted in 32\% (FD002) and 39\% (FD004) relative improvement over the best reported RMSE values (LSTM+Attention). The proposed model also achieved much lower scores for both FD002 and FD004 datasets with an improvement of 82\% and 87\% respectively over the previous best results. Thus, CiRNN is able to reduce the number of late RUL predictions. For FD003 dataset (Table \ref{tab:results3}), the RMSE is at par with other models, however, the score is comparable to only one model (LSTM+FNN) and much higher than the remaining models. In FD001, the CiRNN model performed better than two models, LSTM+FNN and CEEMD+LSTM in terms of RMSE. However, there is a 13\% decrease in performance than the best model (RBM+LSTM). Further, CiRNN score value is higher in comparison to other models. 

From the experimental results, it is apparent that CiRNN models' performance is significantly higher than other models in presence of contexts and comparable to other models when context is not explicitly available (data with one operating condition). Also, in comparison to other models CiRNN achieved the given performance with relatively simpler model with 1 layer and 15 hidden neurons. Feature selection and contextual normalisation also have an influence on the performance of CiRNN.  Further, each of the approaches consider the operating conditions (contextual features) in a different way. For example, Zheng et al. \citep{Zheng2017}, in their approach, clustered the operating conditions and use one-hot encoding for their representation. It is then include as a primary feature. Li et al. \citep{LI2018} used only the sensor data for modeling. In \citep{LISTOUELLEFSEN2019} and \citep{Costa2019}, all the three operational settings and all the sensor measurements are used. However, compared to these approaches,  contextual weighting in CiRNN performs better in all the datasets with explicit contexts.  

\begin{table*}[t!]
\centering
\caption{Comparison of CiRNN performance with existing approaches reported in the literature (FD001 and FD003)}
\label{tab:results2}
\centering
\begin{tabular}{l c c c c }
\hline
\multirow{2}{4em}{Method} &  \multicolumn{2}{c}{FD001} &  \multicolumn{2}{c}{FD003} \\ \cline{2-5}
& RMSE & Score (s)  & RMSE & Score (s) \\
\hline
LSTM+ FNN  & 16.14 & 338 &16.18 & 852\\
\citep{Zheng2017} & & & &\\
CNN + FNN  & 12.61 & 274 &12.64  & 284\\
\citep{LI2018} & & & &\\
RBM + LSTM &\textbf{12.56} &\textbf{231} & \textbf{12.10}& 251\\
\citep{LISTOUELLEFSEN2019}& & & &\\
LSTM + Attention  &13.95& 320 & 12.72 & \textbf{223}\\
\citep{Costa2019}& & & &\\
CEEMD + LSTM  &14.72 &262 &17.72 &452\\
\citep{ZHAO2019}& & & &\\
CiRNN (Proposed) & \textbf{14.53}& \textbf{451} & \textbf{12.87} & \textbf{949}\\
\hline
\end{tabular}
\end{table*}

\begin{table*}[t!]
\renewcommand{\arraystretch}{1.3}
\caption{Comparison of CiRNN performance with existing approaches reported in the literature (FD002 and FD004)}
\label{tab:results3}
\centering
\begin{tabular}{ l c c c c c}
\hline
\multirow{2}{4em}{Method} & \multicolumn{2}{c}{FD002} & \multicolumn{2}{c}{FD004}\\ \cline{2-5}
& RMSE & Score (s) & RMSE & Score (s)\\
\hline
LSTM+ FNN  & 24.49 & 4,450 & 28.17& 5,550\\
\citep{Zheng2017}& & & &\\
CNN + FNN  & 22.36& 10,412 & 23.31& 12,466\\
\citep{LI2018}& & & &\\
RBM + LSTM  & 22.73 & 3,366 &22.66 & 2,840  \\
\citep{LISTOUELLEFSEN2019}& & & &\\
LSTM + Attention & 17.65& 2.102 &20.21 & 3,100 \\
\citep{Costa2019}& & & &\\
CEEMD + LSTM  &29.00 &6953 & 33.43 &15069 \\
\citep{ZHAO2019}& & & &\\
CiRNN (Proposed) & \textbf{11.97} & \textbf{363} & \textbf{12.35} & \textbf{378} \\
\hline
\end{tabular}
\end{table*}

\begin{figure}[tb]
\centering
\includegraphics[width=5.5in, height = 3.5in ]{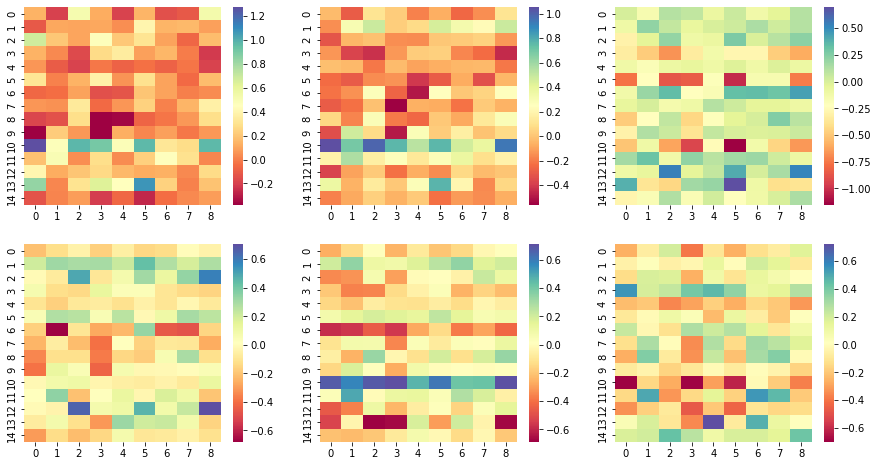}
\caption{Input-to-hidden weights $\mathbf{A}^s$}
\label{fig:weights}
\end{figure}

Finally, in Figure \ref{fig:weights}, the contextual weights $\mathbf{A}^s$ are shown for CiRNN with FD002 dataset (CiRNN\_FD002). Note that this model has 15 hidden units, 6 primary inputs, and 3 contextual inputs. The primary inputs are, $x_1 = s_1$ (total temperature at fan inlet), $x_2 = s_2$ (total temperature at LPC outlet), $x_3 = s_8$ (physical fan speed), $x_4 = s_{13}$ (corrected fan speed), $x_5 = s_{14}$ (corrected core speed), and $x_6 = s_{19}$ (demanded corrected fan speed). With 9 polynomial basis functions of degree 2, the size of $\mathbf{A}^s$ is $(15\times(6\times9))$ i.e. $(15\times54)$. The entire weight matrix is represented through 6 smaller heatmaps of size $15\times9$. The first heatmap represents the weights associated with $x_{1t}\times G(\mathbf{Z}_t)$ and the hidden units. Similarly, second heat map represents the weights associated with   $x_{2t}\times G(\mathbf{Z}_t)$ and the hidden units, and so on. It can be observed from Figure \ref{fig:weights} that weights associated with sensor 1 data (total temperature at fan inlet) , and sensor 2 (total temperature at LPC outlet) are mostly negative. However, hidden unit 10 has connections with positive weights. The connections from sensor 8 data (physical fan speed) has more positive weights compared to remaining 5 sensors. This indicates that physical fan speed has more impact on update gate output and in turn RUL prediction.  

\section{Conclusion}
In this paper, we proposed a novel RNN architecture, named CiRNN,  that enables integrating explicit contexts available from the problem domain. In CiRNN, the input features referred to as primary features are weighted by context which is represented by contextual features. Thus, external factors influences the change in network parameters. To demonstrate the pertinence of CiRNN to applications where contextual information is accessible through sensors, we applied CiRNN to engine health prognostics. In particular, we used CiRNN to estimate the RUL using widely known benchmark dataset (C-MAPSS dataset). The dataset consists of information on operational environment of the flight engines through three settings parameters. This provides the required contextual information for CiRNN. We performed several experiments using datsets with contexts as well as datasets without only one context is available without any variations. The experimental results show that integrating context to RNN (with GRU cell) significantly improves the performance of the resulting CiRNN model in comparison to the baseline models and also the existing approaches with relatively simple network configuration, particularly when context is present (FD002 and FD004). For datasets that do not exhibit context (FD001 and FD003), CiRNN's performance is comparable to the baseline and existing models. 

As our environment is becoming data rich, CiRNN can potentially be applied to many real life application, such as in the domain of smart healthcare, smart manufacturing, and smart agriculture, where contextual information can be acquired and used naturally. In future, we intend to explore such areas. 

\section*{Acknowledgements}
This work is supported by CONEX-Plus programme funded by Universidad Carlos III de Madrid and the European Union’s Horizon 2020 research and innovation programme under the Marie Sklodowska-Curie grant agreement No. 801538.

\appendix

\section{Gradient computation of L2 Loss function with respect to CiRNN parameters}

Let $L_{t}$ be the loss at time step $t$ and $L$ be the total loss over the given time sequence, which are given as,

\begin{equation}
    \begin{split}
    L_{t} = \frac{1}{2}(\mathbf{y}_t - \hat{\mathbf{y}}_t)^2
    \end{split}
\end{equation}

\begin{equation}
    \begin{split}
    L = \frac{1}{2}\sum_t(\mathbf{y}_t - \hat{\mathbf{y}}_t)^2
    \end{split}
\end{equation}

\noindent
To train CiRNN, we need the values of all the parameters $ (\mathbf{A}^h, \mathbf{A}^s, \mathbf{A}^r)$, $(\mathbf{U}^h, \mathbf{U}^s, \mathbf{U}^r)$, and $(\mathbf{V}, \mathbf{b}_y)$ that minimize the loss $L$. For gradient-based optimization, the derivative of $L$ w.r.t. each of the parameters is required. 

Let us first consider ${\partial L}/{\partial \mathbf{A}^s}$. The calculation for the other parameters would be similar. Since $L = 1/2 \sum_t L_t$ and the parameters remain same in each step, we also have $ {\partial L}/{\partial \mathbf{A}^s} = \sum_t (\partial L_t / \partial \mathbf{A}^s)$. So, we can calculate $(\partial L_t / \partial \mathbf{A}^s)$ independently and finally sum them up. 

\begin{equation}
\label{eqn:derv1}
    \begin{split}
        \frac{\partial L_t}{\partial \mathbf{A}^s} & = \frac{\partial L_t}{\partial {\mathbf{h}}_t} \frac{\partial \mathbf{h}_t}{\partial \mathbf{A}^s} \\
    \end{split}
\end{equation}

Let us consider each of the derivatives on the R.H.S. of the equation (\ref{eqn:derv1}).

\begin{equation}
\label{eqn:derv2}
    \begin{split}
        \frac{\partial L_t}{\partial \mathbf{h}_t} & = \frac{\partial L_t}{\partial \hat{\mathbf{y}}_t} \frac{\partial \hat{\mathbf{y}}_t}{\partial \mathbf{h}_t} 
        = (\hat{\mathbf{y}}_t - \mathbf{y}_t) f'(u_t) \mathbf{V} \\
    \end{split}
\end{equation}

\noindent
where $u_t = \mathbf{V} \mathbf{h}_t + \mathbf b_y$

In the second term of R.H.S. of equation (\ref{eqn:derv1}), $\mathbf{h}_t$ also depends on $\mathbf{A}^s$ through $\mathbf{s_t}$ and also depends on $\mathbf{h}_{t-1}$ that in turn depends $\mathbf{A}^s$. So, $\mathbf{h}_t$ depends on $\mathbf{A}^s$ explicitly and also implicitly through $\mathbf{h}_{t-1}$, and we can write $\partial \mathbf{h}_t / \partial \mathbf {A}^s$ as,

\begin{equation}
\label{eqn: exp_imp}
    \begin{split}
        \frac{\partial \mathbf{h}_t}{\partial \mathbf{A}^{s}} =  {\frac{\partial \mathbf{h}_t} {\partial \mathbf{A}^{s}} }^* + \frac{\partial \mathbf{h}_t}{\partial \mathbf{h}_{t-1}}  \frac{\partial \mathbf{h}_{t-1}}{\partial \mathbf{A}^{s}}
    \end{split}
\end{equation}

\noindent
The explicit derivative is denoted by a '$*$' in the above equation where $\mathbf {h}_{t-1}$ is considered as constant. Again, the last term in equation (\ref{eqn: exp_imp}) can be expanded and written in terms of implicit and explicit derivatives as given below:
\begin{equation*}
    \begin{split}
        \frac{\partial \mathbf{h}_t}{\partial \mathbf{A}^s} =  {\frac{\partial \mathbf{h}_t} {\partial \mathbf{A}^s} }^* + \frac{\partial \mathbf{h}_t}{\partial \mathbf{h}_{t-1}}  {\frac{\partial \mathbf{h}_{t-1}}{\partial \mathbf{A}^s}}^* + \frac{\partial \mathbf h_t}{\partial \mathbf h_{t-1}}  \frac{\partial \mathbf h_{t-1}}{\partial \mathbf h_{t-2}} \frac{\partial \mathbf h_{t-2}}{\partial \mathbf{A}^s}
    \end{split}
\end{equation*}

\noindent
This process continues until $\partial \mathbf {h}_1/ \partial \mathbf{A}^s$ is reached. $\mathbf h_1$ is implicitly dependent on $\mathbf h_0$, however, $\mathbf h_0$ is constant and initialized to a vector of zeros. Therefore, equation (\ref{eqn: exp_imp}) can be given as,

\begin{equation}
\label{eqn: exp_imp1}
    \begin{split}
        \frac{\partial \mathbf h_t}{\partial \mathbf{A}^{s}} &=  \sum_{k=1}^t \frac{\partial \mathbf h_t}{\partial \mathbf h_k} {\frac{\partial \mathbf h_k}{\partial \mathbf{A}^s}}^*\\
    \end{split}
\end{equation}

\noindent
where $\frac {\partial \mathbf h_t}{ \partial \mathbf h_k}$ is a chain rule in itself, for example, $\frac {\partial \mathbf h_3}{ \partial \mathbf h_1} = \frac {\partial \mathbf h_3}{ \partial \mathbf h_2}\  \frac {\partial \mathbf h_2}{ \partial \mathbf h_1}$. Further, equation (\ref{eqn: exp_imp1}) can be given as,

\begin{equation}
\label{eqn: exp_imp2}
    \begin{split}
        \frac{\partial \mathbf h_t}{\partial \mathbf{A}^{s}} 
        &= \sum_{k=1}^t \left( \left( \prod_{k=1}^{t-1}\frac{\partial \mathbf h_{k+1}}{\partial \mathbf h_k} \right) {\frac{\partial \mathbf h_k}{\partial \mathbf{A}^s}}^* \right)\\
    \end{split}
\end{equation}

\noindent
Using equations (\ref{eqn:derv1}), (\ref{eqn:derv2}), and (\ref{eqn: exp_imp2}),  ${\partial L_t}/ {\partial \mathbf{A}^s}$ can be given as,

\begin{equation}
    \frac {\partial L_t} {\partial \mathbf{A}^s} = (\hat{\mathbf{y}}_t - \mathbf{y}_t) f'(u_t) \mathbf V \sum_{k=1}^t \left( \left( \prod_{k=1}^{t-1}\frac{\partial \mathbf h_{k+1}}{\partial \mathbf h_k} \right) {\frac{\partial \mathbf h_k}{\partial \mathbf{A}^s}}^* \right)\\
\end{equation}

\noindent
The terms within the summation are derived as follows:

\begin{equation}
    \begin{split}
         {\frac {\partial \mathbf h_t} {\partial \mathbf{A}^s}}^* &= \left( (\mathbf{h}_{t-1} - \tilde{\mathbf{h}}_t) \odot \mathbf{s}_t \odot (1-\mathbf{s}_t) \right)(\mathbf{x}_t \otimes \mathbf{G}(t))^T \\
    \end{split}
\end{equation}

\begin{equation}
    \begin{split}
         {\frac {\partial \mathbf {h}_t} {\partial \mathbf{h}_{t-1}}} &= {\frac{\partial \mathbf{h}_t} {\partial \tilde {\mathbf{h}}_t}} 
         {\frac{\partial \tilde {\mathbf{h}}_t} {\partial \mathbf{h}_{t-1}}} + 
         {\frac{\partial \mathbf{h}_t} {\partial \mathbf{s}_t}} 
         {\frac{\partial \mathbf{s}_t} {\partial \mathbf{h}_{t-1}}}
         + 
         {\frac {\partial \mathbf {h}_t} {\partial \mathbf{h}_{t-1}}}^* \\
         &= {\frac{\partial \mathbf{h}_t} {\partial \tilde {\mathbf{h}}_t}} \left({\frac{\partial \tilde {\mathbf{h}}_t} {\partial \mathbf{r}_{t}}} {\frac{\partial \mathbf{r}_t} {\partial \mathbf{h}_{t-1}}} + {\frac{\partial \tilde {\mathbf{h}}_t} {\partial \mathbf{h}_{t-1}}}^*\right) \\
         &+ {\frac{\partial \mathbf{h}_t} {\partial \mathbf{s}_{t}}} {\frac{\partial \mathbf{s}_t} {\partial \mathbf{h}_{t-1}}} + {\frac{\partial \mathbf{h}_t} {\partial \mathbf{h}_{t-1}}}^*
    \end{split}
\end{equation}

\begin{equation}
    \begin{split}
         {\frac {\partial \mathbf {h}_t} {\partial \mathbf{h}_{t-1}}} &= (1-\mathbf{s}_t)T_1 + T_2 + \mathbf{s}_t \\
         T_1 &= (\mathbf{U}^{rT}((\mathbf{U}^{hT}(1-\tilde{\mathbf{h}}_t \odot \tilde{\mathbf{h}}_t)) \odot \mathbf{h}_{t-1} \odot \mathbf{r}_t \\
         & \odot (1-\mathbf{r}_t))+ ((\mathbf{U}^{hT}(1-\tilde{\mathbf{h}}_t \odot \tilde{\mathbf{h}}_t)) \odot \mathbf{r}_t ) \\
         T_2 &= \mathbf{U}^{sT}((\mathbf{h}_{t-1} - \tilde {\mathbf{h}}_t) \odot \mathbf{s}_t \odot (1-\mathbf{s}_t))\\
    \end{split}
\end{equation}

So far, we have determined all the components required for $\partial L_t / \partial \mathbf{A}^s$. The gradient of $L_t$ with respect to other parameters are similar.

\bibliographystyle{unsrtnat}
\bibliography{references}  






\end{document}